\pdfoutput=1

\documentclass[11pt]{article}

\usepackage[preprint]{acl}

\usepackage{times}
\usepackage{latexsym}

\usepackage[T1]{fontenc}

\usepackage[utf8]{inputenc}

\usepackage{microtype}

\usepackage{inconsolata}

\usepackage{graphicx}

\usepackage{booktabs}

\usepackage{colortbl}
\usepackage{multirow}
\usepackage{pifont}
\usepackage{listings}

\usepackage{caption}

\usepackage{amsmath}
\usepackage{cleveref}
\usepackage[leftcaption]{sidecap}

\definecolor{backcolor}{rgb}{0.95,0.95,0.92}
\definecolor{LightCyan}{rgb}{0.88,1,1}
\definecolor{emerald}{RGB}{80,200,120}
\definecolor{aliceblue}{rgb}{0.94,0.97,1.0}

\newcommand{\green}[1]{\textcolor{emerald}{#1}}
\newcommand{\vi}{$\mathcal{VI}$ }

\newcommand{\cmark}{\ding{51}}%
\newcommand{\xmark}{\ding{55}}

\NewDocumentCommand{\xudong}{ mO{} }{\textcolor{blue}{\textsuperscript{\textit{Xudong}}\textsf{\textbf{\small[#1]}}}}

\usepackage[accsupp]{axessibility}  

    \title{VIEWS: Entity-Aware News Video Captioning}

\author{
 \textbf{Hammad Ayyubi\textsuperscript{1}},\quad
 \textbf{Tianqi Liu\textsuperscript{2}},\quad
 \textbf{Arsha Nagrani\textsuperscript{2}},\quad
 \textbf{Xudong Lin\textsuperscript{1}},\quad
  \\[3pt]
 \textbf{Mingda Zhang\textsuperscript{2}},\quad
 \textbf{Anurag Arnab\textsuperscript{2}},\quad
 \textbf{Feng Han\textsuperscript{2}},\quad
 \textbf{Yukun Zhu \textsuperscript{2}},\quad
  \\[3pt]
 \textbf{Xuande Feng\textsuperscript{1}},\quad
 \textbf{Kevin Zhang\textsuperscript{1}},\quad
 \textbf{Jialu Liu\textsuperscript{2}},\quad
 \textbf{Shih-Fu Chang\textsuperscript{1}},
\\[4pt]
 \textsuperscript{1}Columbia University,\quad
 \textsuperscript{2}Google,\quad
\\
 \small{
   \textbf{Correspondence:} \href{hayyubi@cs.columbia.edu}{hayyubi@cs.columbia.edu}
 }
}

\begin{document}
\maketitle

\begin{abstract}
Existing popular video captioning benchmarks and models often produce generic captions for videos that lack specific identification of individuals, locations, or organizations (named-entities). However, in the case of news videos, the setting is more demanding, requiring the inclusion of such named entities for meaningful summarization. Therefore, we introduce the task of directly summarizing news videos into captions that are entities-aware. To facilitate research in this area, we have collected a large-scale dataset named VIEWS (VIdeo NEWS). Within this task, we face challenges inherent to recognizing named entities and navigating diverse, dynamic contexts, all while relying solely on visual cues. To address these challenges, we propose a model-agnostic approach that enriches visual information extracted from videos with context sourced from external knowledge, enabling the generation of entity-aware captions. We validate the effectiveness of our approach across three video captioning models. Additionally, we conduct a critical analysis of our methodology to gain insights into the complexity of the task, the challenges it presents, and potential avenues for future research.
\end{abstract}    
\section{Introduction}
\label{sec:intro}

\begin{figure}[h]
\captionsetup{skip=3pt}
\centering
  \includegraphics[width=\linewidth]{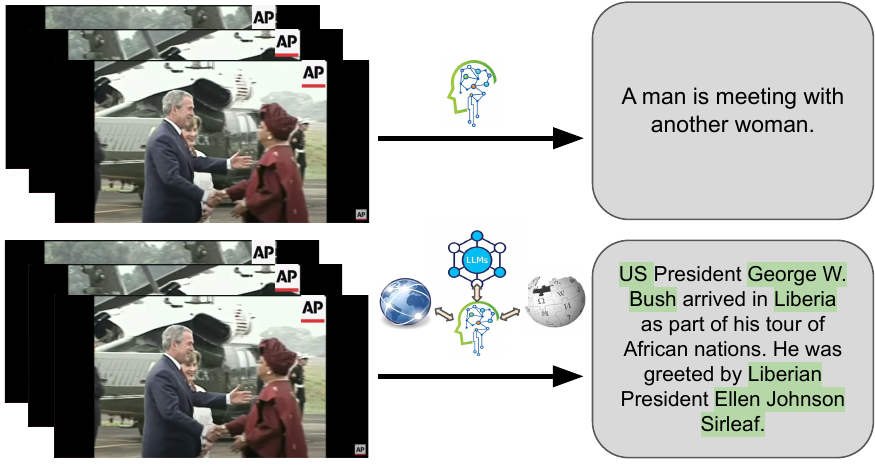}
  \caption{Top: In the traditional Video Captioning task, captions typically do not contain specific named-entities.
Bottom: In the proposed Entity-Aware Video Captioning task, the objective is to generate captions that include relevant entities, such as highlighted names and places. This is necessary for effective captioning in the news domain. The task permits utilization of external knowledge sources.}
  \label{fig:intro}
\end{figure}

Summarizing video content into a natural language description (i.e. `video captioning') is a crucial aspect of video understanding. It typically requires understanding salient information and dynamics displayed in the video. Existing datasets typically describe videos using largely generic captions: ``A woman is giving a speech on stage" (MSR-VTT \cite{Xu2016MSRVTTAL}), ``An elderly man is playing the piano
in front of a crowd" (ActivityNet Captions \cite{krishna2017densecaptioning}), ``pour in spicy sauce" (YouCook2 \cite{ZhXuCoAAAI18}) etc. While these captions suffice for daily activity or cooking videos, they are rendered almost meaningless in the domain of news videos. As shown in \cref{fig:intro}, a generic caption saying ``a man meets another woman'' provides little to no information about: 1) Who is meeting whom? 2) Where? 3) In what context? 4) In relation to which organization? In other words, such captions lack \textit{named entities}.

\begin{figure*}[t]
\centering
\captionsetup{skip=3pt}
  \includegraphics[width=\linewidth]{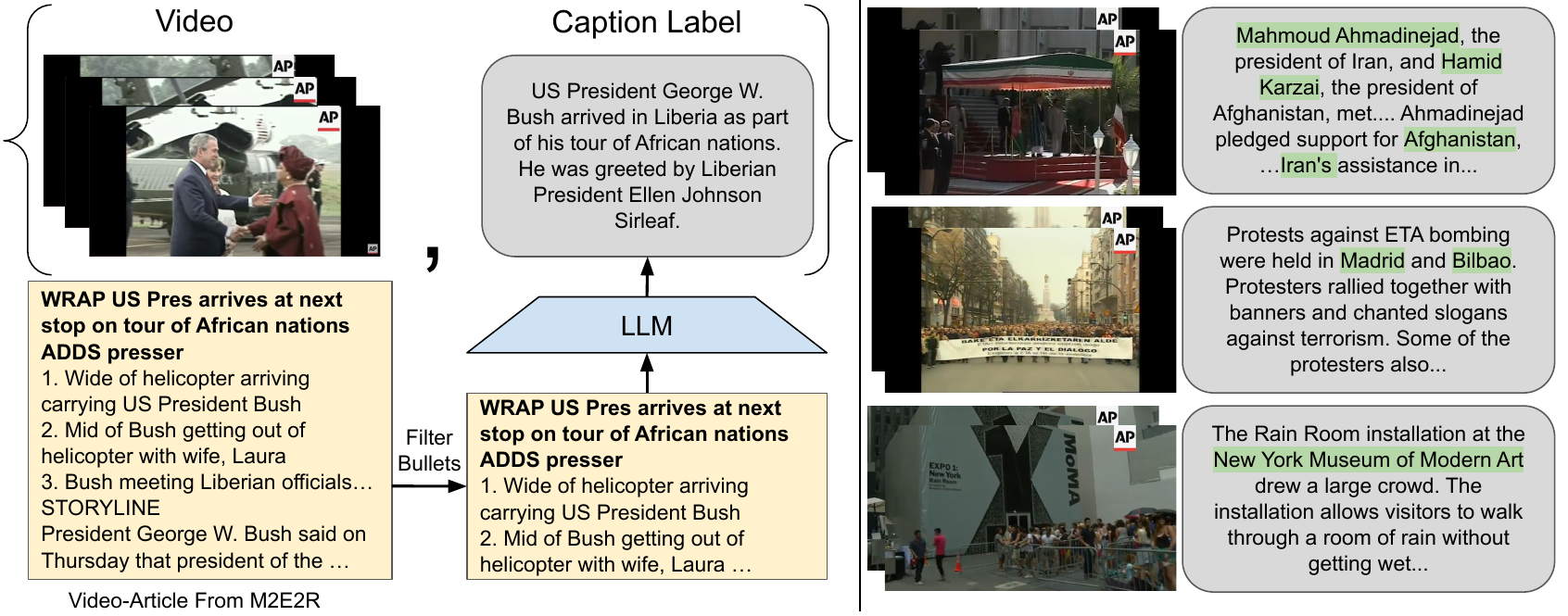}
  \caption{Left: Data Collection Procedure. Right: Samples from our dataset highlighting key properties: captions are rich in entities (colored green), well aligned with the video and diverse (topics shown: politics, conflict \& art).}
  \label{fig:data}
\end{figure*}

Indeed, summarizing news videos into entity-aware captions is a challenging task. There has been a line of recent works which additionally utilize the news article paired with the video to generate entity aware captions \cite{krubinski-pecina-2023-mlask, fu-etal-2021-mm, liu2021visual}. However, this assumes that a paired news article will always be available for the video. We argue that this is an impractical assumption -- there are a number of standalone news video content uploaded online, especially by individuals uninvolved with big news media corporations.


In this paper, we introduce the task of summarizing news videos into entity-aware captions. To support research on this task, we present VIdeo NEWS (VIEWS\footnote{Data: \url{https://github.com/hayyubi/views}}), comprising a large-scale collection of news videos accompanied by entity-rich captions, totaling 144K samples. Unlike previous news video datasets, we rigorously verify and ensure high alignment between the captions and the corresponding videos. Approximately 94\% of samples in our training set exhibit high precision, with any remaining precision or alignment issues in the test set manually corrected. Upon acceptance, we will publicly release the dataset.

This task and dataset introduces numerous challenges. Firstly, it requires models to explicitly use visual cues to recognize people, places, events, organizations etc. (named entities) depicted in the video. For instance, in \cref{fig:intro}, it is required to recognize the people involved in the meeting -- George Bush and Ellen Johnson. The landmarks and cultural clues need to be utilized to predict that the meeting is taking place in Liberia. Secondly, understanding the background context from the presented video is a non-trivial task. In \cref{fig:intro}, this would mean identifying that George Bush is not just touring Liberia, instead, this meeting is part of his African Nations tour. Further, context is incredibly hard to learn -- it's almost unique for every new sample and even changes for the same entities.
Lastly, these entities and contexts need to be seamlessly incorporated in the final captioning process.

We propose a novel approach to tackle these challenges. The approach first trains an Entity-Perceiver(EP) to recognize named entities directly from the video. Then, we feed the detected entities to a Knowledge-Extractor (KE) to extract a relevant news data as context from external source. Finally, we leverage recent transformer-based video captioning models to integrate the video, the entities, and the context into informative entity-aware captions. We demonstrate the effectiveness of our approach on three state-of-the-art models and study in detail the impact of each of our modules. A detailed analysis leads to many insights and conclusions on where the complexity of the task comes from and how to address them in future works.
\section{Related Work}
\label{sec:related_work}

\begin{table}[t]
\captionsetup{skip=3pt}
\centering
\fontsize{8}{10}\selectfont
\setlength{\tabcolsep}{2pt}

\resizebox{1.\linewidth}{!}{
\begin{tabular}{l l l c c}
\toprule

\multirow{2}*{Dataset}       & \multirow{2}*{Domain}  & \multirow{2}*{\#Videos}  & Entities & Cap.-Vid \\
&&&  in Cap. & Align $\uparrow$ \\
\midrule
MSR-VTT~\cite{Xu2016MSRVTTAL}       & Open    & 7.18K          & \xmark    & \cmark \\
ActyNet Cap~\cite{krishna2017densecaptioning}   & Open    & 20K         & \xmark  & \cmark    \\
YouCook II~\cite{ZhXuCoAAAI18}    & Cooking & 2K           & \xmark   & \cmark  \\
TVR/TVC~\cite{lei2020tvr}       & TV show & 21.8K     & \xmark *   & \cmark  \\
LSMDC~\cite{rohrbach2016movie}         & Movie   & 128K     & \xmark *  & \cmark  \\
MM-AVS~\cite{fu-etal-2021-mm}        & News    & 2.17K  & \cmark    &  \xmark  \\
MLASK~\cite{krubinski-pecina-2023-mlask}         & News    & 41.2K     & \cmark  &   \xmark  \\

\midrule
VIEWS (Ours)  & News    & 144K      & \cmark  & \cmark \\
\bottomrule
\end{tabular}
}
\caption{In comparison to other Video Captioning datasets, VIEWS is the only one to contain entity-rich captions and highly video-aligned captions. Cap: Captions. Vid: Video. s: seconds. \xmark*: grounds entities (Person 1/2/3) but doesn't name them.}
\label{tab:dataset-comp}
\end{table}

\textbf{Video Captioning.} Video captioning is a long-standing task that has been explored in various domains: daily activities \cite{chen2011collecting, Xu2016MSRVTTAL, krishna2017densecaptioning}, cooking \cite{ZhXuCoAAAI18, rohrbach15ijcv} and movies/TV shows \cite{bain2020condensed, lei2020tvr, rohrbach2016movie}. 
However, these established video captioning benchmarks do not cover named entities in the real world. 
This leads to a vacancy in model development as well: recent multimodal Large Language Models (LLMs) \cite{alayrac2022flamingo} demonstrate impressive performance on traditional benchmarks, but tend to hallucinate named-entities \cite{ji2023survey, pope}.
We address this gap through the proposed dataset, VIEWS.
VIEWS contains captions that are rich in entities by focusing on the news domain, and encourages entities-aware model development.

\noindent \textbf{News Content Understanding.}
Recent works have attempted to specifically generate entity-aware captions for videos \cite{fu-etal-2021-mm, tang2022tldw, liang-etal-2023-summary, whitehead-etal-2018-incorporating} and images \cite{biten2019good, Tran_2020_CVPR, lu-etal-2018-entity, zhao-etal-2019-informative}.
However, they require paired news articles or additional metadata (tags, dates, or named-entities) to do so.
This makes these approaches unrealistic in real life, where paired articles or additional metadata may not be readily available.
Further, it's not easily scalable.
We address these limitations via the novel task of summarizing news videos directly to entity-aware captions and establishing a large-scale dataset specifically for training and evaluation.


\noindent \textbf{External Knowledge.}
To acquire information beyond the training set, many approaches leverage external knowledge sources such as Wikimedia \cite{chen2021krit}, Google Search \cite{lin2022retrieval}, Wikidata \cite{hu2023reveal}, and so on.
Orthogonally, \citet{yu2023generate} argue that using an LLM to generate knowledge is better than retrieving it. 
\citet{zhang2023multicot} extend this idea to images.
Building on these works, we use LLMs as a knowledge source for videos. Unlike previous approaches, we propose a two-stage pipeline that combines learning and generation for more factual knowledge.

\section{Dataset}
\label{sec:dataset}

\begin{figure}[t]
    \centering
    \captionsetup{skip=-15pt}
    \includegraphics[width=0.85\linewidth]{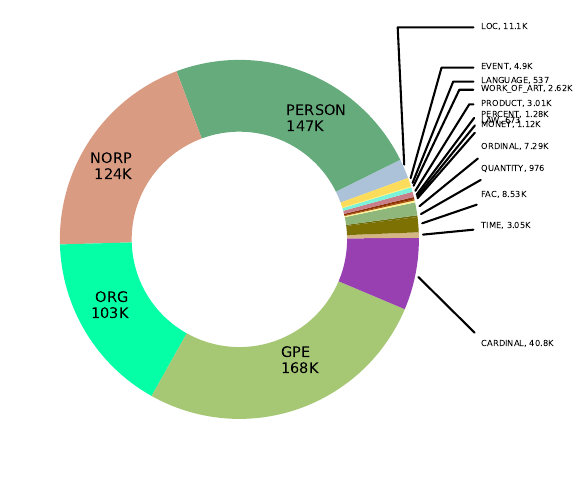}
    \caption{Distribution of entities in VIEWS. NORP: Nationalities or religious or political groups. GPE: geopolitical entities (countries, cities, states). ORG: organizations (companies, agencies, institutions).}
    \label{fig:entities_pie}
\end{figure}

\begin{figure}[t]
\captionsetup{skip=3pt}
    \centering
    \includegraphics[width=0.85\linewidth]{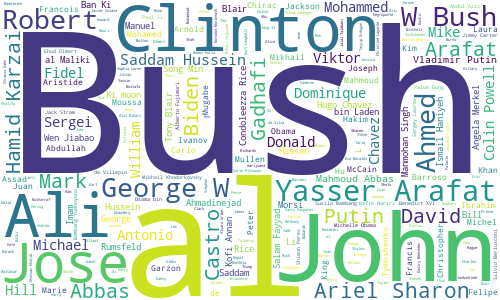}
    \caption{Word cloud of `PERSON' entities in VIEWS.}
    \label{fig:entities_person}
\end{figure}

In this section we describe our newly collected  VIdeo NEWS (VIEWS) dataset which contains news videos accompanied by captions and additional metadata such as the title, a paired news article and the news media source. We list further details below.

\noindent \textbf{Dataset Collection.} We build our dataset on top of an existing news video-article dataset, M$^2$E$^2$R \cite{ayyubi2022multimodal}. 
M$^2$E$^2$R has a total of 220K videos-article pairs sourced from diverse news media outlets like Associated Press, BBC News, Axios, NPR etc. We leverage the articles paired with videos to generate caption for the videos. 
A straightforward approach would be to generate a summary from the article. However, articles often contain related historical stories, opinions, and predictions in addition to describing the video's content. Including such information in the caption would render it unfaithful to the video, as this information is not visually depicted.

To mitigate this issue, we exploit an interesting property of M$^2$E$^2$R -- some articles contain an explicit event description of each camera shot in the video, apart from the regular article story. We leverage these event descriptions and title (for context) to generate a video caption using a LLM (cf. \Cref{fig:data}). 
Specifically, we use PaLM-2-L-IT \cite{anil2023palm} (more details in Appendix). This results in captions that are highly aligned with video content. 
The filtering step to collect videos having event descriptions in article leads to a total data size of $\sim$144K. We split this data into 1.6K dev and test set each, leaving $~$141K for training. 


\begin{figure*}[t]
\centering
\captionsetup{skip=3pt}
  \includegraphics[width=\linewidth]{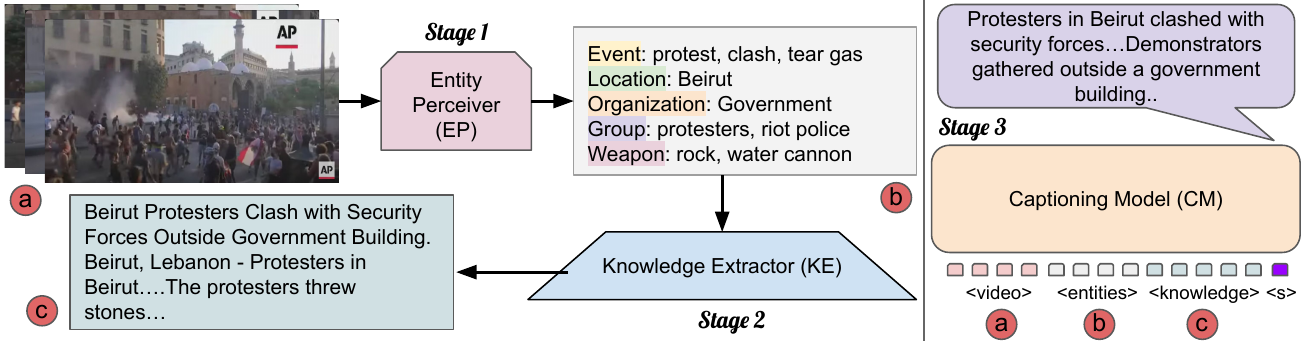}
  \caption{An overview of our proposed approach. Given the video input, we first use EP to detect entities. Then, KE uses the detected entities to extract contextual knowledge about the video. Finally, video, entities and context are input to CM to generate entity-aware video captions.}
  \label{fig:method}
\end{figure*}

\noindent \textbf{Quality Control and Annotation.} While the captions are already highly aligned with the video, we do further quality control, verification and rectification on dev/test set.
We employ a human-in-the-loop procedure for this task.
First, we ask a LLM rater
to score the quality of captions according to the following criteria: 1) It should contain the name of important entities (people, place, organizations etc.); 2) It should not hallucinate 
; 3) It should not leave out critical information from the event descriptions. The samples that the rater flagged as not being of high quality are separated for manual verification and correction. 
For rating, we used LLaMA-2 \cite{touvron2023llama}. 

Separately, we analysed how well the strategy of using LLM raters aligned with human judgement of a high quality caption. For this purpose, we manually rated a gold set of 100 samples and also asked the LLM raters to rate it. We found that only $\sim$6\% times we flagged a caption to be of poor quality and LLM raters failed to flag it. 

\noindent \textbf{Comparison with Existing Datasets.} As reported in \Cref{tab:dataset-comp}, our VIEWS dataset significantly increases the scale of the news video captioning dataset by one to two orders of magnitude. 
Moreover, our VIEWS dataset is the first news video captioning dataset that has a high alignment between captions and videos as other datasets' captions are derived from paired articles.
This establishes the basis for training and evaluation of models generating captions from video content without access to paired articles.
Compared to existing video datasets in other domains, our VIEWS dataset is focused on entity-aware captions, which cover numerous real-world entities, including people, places, events, organizations, etc. (see \Cref{fig:entities_pie} \& \Cref{fig:entities_person}). The large-scale nature, the high video-caption alignment, and the focus on entity-ware caption make VIEWS first-of-its-kind in the community, to the best of our knowledge.
We provide more dataset curation details, statistics, comparisons, and samples in \Cref{sec:dataset_details}.

\section{Method}
\label{sec:method}

Summarizing news videos directly to entity-aware captions is a challenging task. It requires not just recognition of named entities from people, places, and landmarks, but also an understanding of the context in which the events in the video are unfolding. 
The straightforward way of directly training the model with supervision from entity-aware ground-truth captions may enable recognition of frequently occurring named entities. However, learning context this way is nearly impossible since context changes dynamically with every sample. 

To address this, we divide the learning into three stages: 1) Entity Perceiver (EP) is trained to detect high-frequency entities from videos. 2) Knowledge Extractor (KE) uses detected entities to extract relevant context from external sources, 
eliminating the need for learning contexts. 
The extracted context also often contains missing long-tailed entities. 3) Captioning Model incorporates video, entities and context into informative entity-aware captions. The method is depicted visually in \cref{fig:method}. We detail each of the components below.

\noindent \textbf{Entity Perciever (EP).} The objective of EP is to recognize entities in the given video. Unlike in NLP, there doesn't exist open-domain off-the-shelf entity detector from vision data. Consequently, we train our own Entity Perciever (EP). We use a transformer based sequence-to-sequence generative model instead of a discriminative model for EP. This choice  is motivated by the fact that for open-domain entities 
the number of entity types (`Person', `Location' etc.) and their instantiations may vary significantly across samples as shown in \Cref{fig:entities-comp}. 
A generative model facilitates this property seamlessly.
EP inputs videos and outputs entities relating each entity with its type in a dictionary format, as shown in \cref{fig:method}, Stage 1.
To train EP, we extract ground-truth entities from the given caption label using a LLM. 
More details on this process in \Cref{sec:method_details}.

Formally, we train EP $\mathcal{F}_{EP}: \mathbb{R}^{H \times W \times 3 \times F} \longrightarrow \mathbb{R}^{N \times L_E}$ 
using a language modelling objective \cite{sutskever2014sequence} following~\cite{wang2022git} with this loss,
\begin{equation*}
\resizebox{1.\hsize}{!}{$
    \mathcal{L}_{EP} = \frac{1}{L_E} \sum_i \text{CE}( \mathbf{e}_i,  \mathcal{F}_{EP} (\mathbf{v}, \mathbf{e}_{1},..., \mathbf{e}_{i-1}))
    $}
\end{equation*}
where 
$L_E$ is the number of tokens in the target sequence, $\text{CE}$ stands for the standard cross-entropy loss, $\mathbf{v} \in \mathbb{R}^{H \times W \times 3 \times F}$ is the input video, and $\mathbf{e}_{i}$ denotes the $i$-th ground-truth token in the target sequence. Note that this process is done separately before the captioning model is trained.

\noindent \textbf{Knowledge Extractor (KE).} KE is required to reason about the event in the video given entities and extract a relevant news data as context from an external knowledge source (depicted in \Cref{fig:method}, Stage 2). In this work, we use a LLM for this purpose. LLMs are typically trained on massive corpus of internet text, which includes news data. This coupled with the LLM’s reasoning abilities allows extraction of relevant news for a given entities query.
In contrast, KE sources such as Wikipedia \cite{hu2023reveal} or Encyclopedia \cite{mensink2023encyclopedic} often don’t cover lesser known world affairs such as Bush’s tour of Africa Nation in \Cref{fig:intro} or Cyril Svoboda and Condoleeza Rice’s meeting in \Cref{fig:entities-comp}. Further, KE sources such as web searches lack LLM’s reasoning abilities and are quite sensitive to search query. Besides, there are limited public APIs
available for web searches at this time. 

Formally, this process entails generation of context sequence $\mathbf{k}_{1},..., \mathbf{k}_{L_K}$ by KE $\mathcal{F}_{KE}$, given a prompt sequence $\mathbf{p}_{1},..., \mathbf{p}_{L_P}$ and detected entities $\hat{\mathbf{e}}_{1},..., \hat{\mathbf{e}}_{L_E}$. As shown in the equation:
\begin{equation*}
\resizebox{0.9\hsize}{!}{$
    \mathbf{k}_{1},..., \mathbf{k}_{L_K} = \mathcal{F}_{KE} (\mathbf{p}_{1},..., \mathbf{p}_{L_P}, \hat{\mathbf{e}}_{1},..., \hat{\mathbf{e}}_{L_E}),
    $}
\end{equation*}
We detail the used prompt in \Cref{sec:method_details}.

Notably, our approach keeps the LLM as an independent module and allows it to be replaced by any other sources of KE or even new/updated LLMs in future. 
We show how this enables our method to adapt to novel news content in \Cref{sec:exp_generalization}.


\noindent \textbf{Captioning Model (CM).} Henceforth, we refer to the detected entities and extracted context together as Video Information ($\mathcal{VI}$). CM integrates the video with \vi to generate entity and context aware video caption (illustrated in \Cref{fig:method}, Stage 3). 
This is achieved via inbuilt transformer based attention mechanism inspired by \cite{lewis2020retrieval, mensink2023encyclopedic}. 
We demonstrate the effectiveness of our approach on three state-of-the-art CMs (more details in \Cref{sec:exp_baselines}).
It is important to note the versatility of our approach: it can be applied to any standard multimodal CM.

Despite the models' different detailed architecture design, the training loss can be formalized as:
\begin{equation*} 
\resizebox{0.8\hsize}{!}{$
\begin{split}
    \mathcal{L}_{CM} = \frac{1}{L_C} \sum_i \text{CE}( \mathbf{h}_i, \mathcal{F}_{Cap} (\mathbf{v}, \hat{\mathbf{e}}_{1},..., \hat{\mathbf{e}}_{L_E},\\
    \mathbf{k}_{1},..., \mathbf{k}_{L_K}, \mathbf{h}_{1},..., \mathbf{h}_{i-1}))
    \end{split}
$}
\end{equation*}
where $L_C$ is the length of caption, $\mathcal{F}_{Cap}$ is the captioning model, and $\mathbf{h}_i$ is $i$-th token in the ground-truth caption.

\section{Experiments}
\label{sec:experiments} 


\begin{table}[t]
\centering
\captionsetup{skip=3pt}
\fontsize{9}{11}\selectfont
\setlength{\tabcolsep}{4pt}
\begin{tabular}{l|lll|l}
\toprule
Model  & B-4   & R-L & CIDEr & Ent. F1 \\
\midrule
\color{gray} GIT  & \color{gray}  0.06 & \color{gray} 3.99  & \color{gray} 0.01  &\color{gray} 0.75 \\
\color{gray} Video-LLaMA & \color{gray} 0.40 & \color{gray} 12.90 & \color{gray} 0.80 & \color{gray} 1.45 \\

\midrule
GIT  & 5.40 & 22.06  & 18.04 & 16.20 \\
\rowcolor{aliceblue}
\hspace{5pt} + $\mathcal{VI}$ \small{(Ours)}  & {5.80} & {22.45} & {22.27}$_{\green{\Delta}{4.23}}$ & {17.89}$_{\green{\Delta}1.69}$\\
\midrule
BLIP-2  & 4.75 & 22.28 & 13.38 & 28.97 \\
\rowcolor{aliceblue}
\hspace{5pt} + $\mathcal{VI}$ \small{(Ours)}  & {4.89} & {22.31}  & {14.41}$_{\green{\Delta}1.03}$ & 29.11$_{\green{\Delta}0.14}$\\
\midrule
ViT-T5  & 3.05 & 18.29 &  7.24 & 19.20 \\
\rowcolor{aliceblue}
\hspace{5pt} + $\mathcal{VI}$ \small{(Ours)} & 3.48 & 18.92  & 9.87$_{\green{\Delta}2.63}$  &  22.77$_{\green{\Delta}3.57}$ \\
\bottomrule
\end{tabular}
\caption{The impact of using our proposed approach (+$\mathcal{VI}$) on generating entity-aware captions in VIEWS dataset. 
Our approach improves the performance of all models across all metrics.
{\color{gray} Greyed rows} are evaluated in a zero-shot manner. 
}
\label{tab:main-result}
\end{table}

\subsection{Baselines}
\label{sec:exp_baselines}
We carefully choose three baseline Captioning Models (CMs) to ensure we can analyse our approach on diverse paradigms of architectures/training strategies.

\noindent \textbf{GIT}\cite{wang2022git} represents the class of traditional encoder-decoder model. 
We use the official pretrained GIT-L and finetune the model end-to-end with \#frames=19, \#maximum caption tokens=100, and \#maximum $\mathcal{VI}$ tokens=300.

\noindent \textbf{BLIP-2}\cite{li2023blip2} represents the category of models that exploit the power of (frozen) LLM via Q-Former \cite{li2023blip2} to generate captions from visual data.
We finetune the official BLIP-2 pretrained weights with
13 \#frames, the maximum we can fit in the memory. The maximum caption and $\mathcal{VI}$ tokens remain the same as GIT.

\noindent \textbf{ViT-T5.} GIT trains the model end-to-end but uses a relatively small decoder, while BLIP-2 uses strong LLM decoder but keeps it frozen. As such, we construct this baseline with a large LLM decoder (T5-L \cite{chung2022scaling}) and finetune it end-to-end. 
The rest of the implementation details remain the same as BLIP-2.

Additional baseline implementation details are provided in \Cref{sec:baselines_details}. We report the main results on the test set. All ablations and analyses are reported on the validation set using GIT, unless otherwise stated.

\begin{figure}[t]
\captionsetup{skip=-3pt}
\centering
    \includegraphics[width=\linewidth]{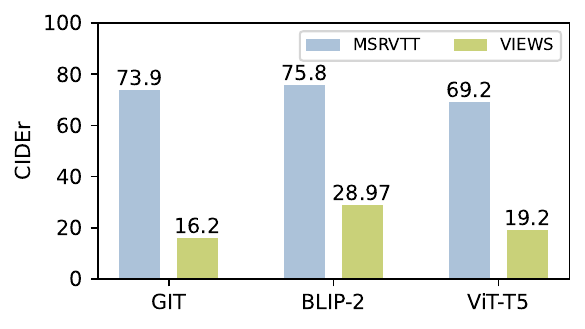}
    \caption{Traditional captioning (MSRVTT) vs Entity-aware captioning (VIEWS). Performance drop $>$60\% on VIEWS across all models highlights its complexity.}
    \label{fig:msrvtt_vs_views}
\end{figure}

\subsection{Experimental Setup}
\label{sec:exp_setup}
\textbf{Implementation Details.} For the Entity Perceiver (EP), we use a pre-trained GIT model and for Knowledge Extractor (KE), we use PaLM-2-L-IT LLM\cite{anil2023palm}. 

\noindent \textbf{Training Details.} We train GIT, BLIP-2 and VIT-T5 for 42, 40 and 30 epochs respectively, using a batch size of 128, a learning rate of 1e-5, adamW  \cite{loshchilov2017decoupled} optimizer and cosine scheduler with a linear warmup. 

\noindent\textbf{Metrics.} We use the standard BLEU-4 (B-4), Rouge-L (R-L) and CIDEr metrics used in captioning tasks \cite{vedantam2015cider, lin2004rouge, papineni2002bleu} . We also calculate an Entity F$_1$ (Ent. F$_1$) score to measure how well the ground-truth entities are covered in the generated caption, similar to \cite{liu2021visual, bertsch2023unlimiformer}. To this end, named-entities are extracted from ground-truth caption and generated caption, and then compared using F$_1$ score. Among all metrics, we emphasize CIDEr and Ent F$_1$ the most. CIDEr is important for evaluating caption quality as it strikes a balance between precision focused B-4 and recall focused R-L. On the other hand, Ent F$_1$ is crucial for evaluating entities presence in the captions.


 

\begin{SCtable}[1.7][t]
\centering
\fontsize{9}{11}\selectfont
\setlength{\tabcolsep}{4pt}
\caption{If available, video-paired article significantly boosts model performance. This implies unavailability of articles in VIEWS makes it challenging.}
\hspace{-6pt}
\begin{tabular}{l c}
\toprule
Method    &     CIDEr\\
\midrule
GIT &  17.78 \\
\hspace{3pt} + Article & \textbf{22.20} \\
\bottomrule
\end{tabular}
\vspace{-3pt}
\label{tab:article_addition}
\end{SCtable}
\subsection{Main Results: Proposed Method's Impact}
We study how incorporating entities and context ($\mathcal{VI}$) to the videos via our proposed approach (+ $\mathcal{VI}$) improves CMs' performance on the task of generating entity-aware captions. The results are reported in \Cref{tab:main-result}. We make the following observation:

\noindent \textbf{Our method improves all three captioning models' performance.} For all three models, our approach registers improvements across all the metrics with significant increments in CIDEr (Avg. \textasciitilde 22.5\%) and Ent. F$_1$ (Agv. \textasciitilde 10\%). This demonstrates our approach's effectiveness in infusing existing CMs with the ability to generate context and entity aware captions. Moreover, it signifies our method's model-agnostic behavior.

\noindent \textbf{End-to-end finetuned models improve most.} Avg. improvement for finetuned models (GIT and ViT-T5) is much larger when compared to partially-tuned BLIP-2 (30\% vs 8\% CIDEr and 14.5\% vs 0.5\% Ent. F$_1$). This indicates end-to-end training is best suited to adapt models to utilize our $\mathcal{VI}$.

\noindent \textbf{Zero-shot models perform poorly.} Git and even LLM based, Video-LLaMA, perform quite poorly on this task. This highlights the complexity of the task and exposes existing model's inability to generate entity-aware captions.


\begin{table}[t]
\centering
\captionsetup{skip=3pt}
\fontsize{9}{11}\selectfont
\begin{tabular}{llc}
\toprule





& {Ablation} & CIDEr      \\
\midrule
1. & Ours &  \textbf{21.00}   \\

2. & \hspace{5pt} - w/o Entities & 20.31 \\

3. & \hspace{5pt} - w/o Knowledge  & 20.75 \\

4. & \hspace{5pt} - w/o \vi & 17.78  \\

\bottomrule
\end{tabular}
\captionof{table}{Ablation study on GIT to study the impact of our design choices. Both entities and context positively impact the model's performance.}
\label{tab:ablations-component}
\end{table}
\begin{table}[t]
\centering
\captionsetup{skip=6pt}
\fontsize{9}{11}\selectfont
\setlength{\tabcolsep}{4pt}
\begin{tabular}{l c}
\toprule
GT Caption Source    &     CIDEr\\
\midrule
Event Descriptions &  \textbf{18.04} \\
Paired Article & 10.5 \\
\bottomrule
\end{tabular}
\caption{Ground-Truth (GT) captions sourced from articles perform poorly compared to captions from event descriptions. This indicates captions derived from articles are not well aligned with the video}
\label{tab:captions-from-articles}
\end{table}
\subsection{Analysis of Task Complexity}
The low zero-shot and even finetuned baseline performance on VIEWS hints at high task complexity. In this section, we analyze various factors that make it so.

\noindent\textbf{Entity-aware video captioning is challenging.} In \Cref{fig:msrvtt_vs_views}, we compare CIDEr scores of three baselines on a traditional captioning benchmark, MSRVTT, against our entity-aware benchmark, VIEWS. We find that all the baselines undergo at least 61\% performance drop from MSRVTT to VIEWS. This emphasizes the complexity of our task.

\noindent \textbf{Unavailability of video paired article makes VIEWS challenging.} VIEWS requires generating entity-aware captions directly from videos to be applicable in practical settings where video-paired news article is unavailable. 
If available, we find that it leads to performance boost of 24.85\% in table \Cref{tab:article_addition}. This illustrates that part of VIEWS's complexity comes from unavailability of paired article.

\noindent \textbf{Captions derived from paired article are not well aligned with the video and thus noisy.}
To justify our usage of event descriptions instead of paired articles to generate ground-truth video captions, we train the model on captions generated from event descriptions and paired articles respectively. We report the results in \Cref{tab:captions-from-articles}. We find that training the model on article-sourced-captions leads to a drop of 71.81\% performance as compared to event-descriptions-sourced-captions. This demonstrates that the paired article, and hence article-sourced-caption, has a noisy relationship with the visual content.

\subsection{Analysis of Proposed Method}
\label{sec:ablation}
The proposed method relies heavily on detected entities and extracted context. As such, we evaluate their quality and impact on model performance.

\begin{figure}[t]
\captionsetup{skip=3pt}
\centering
  \begin{minipage}[t]{\linewidth}
    \centering
      \includegraphics[width=\linewidth]{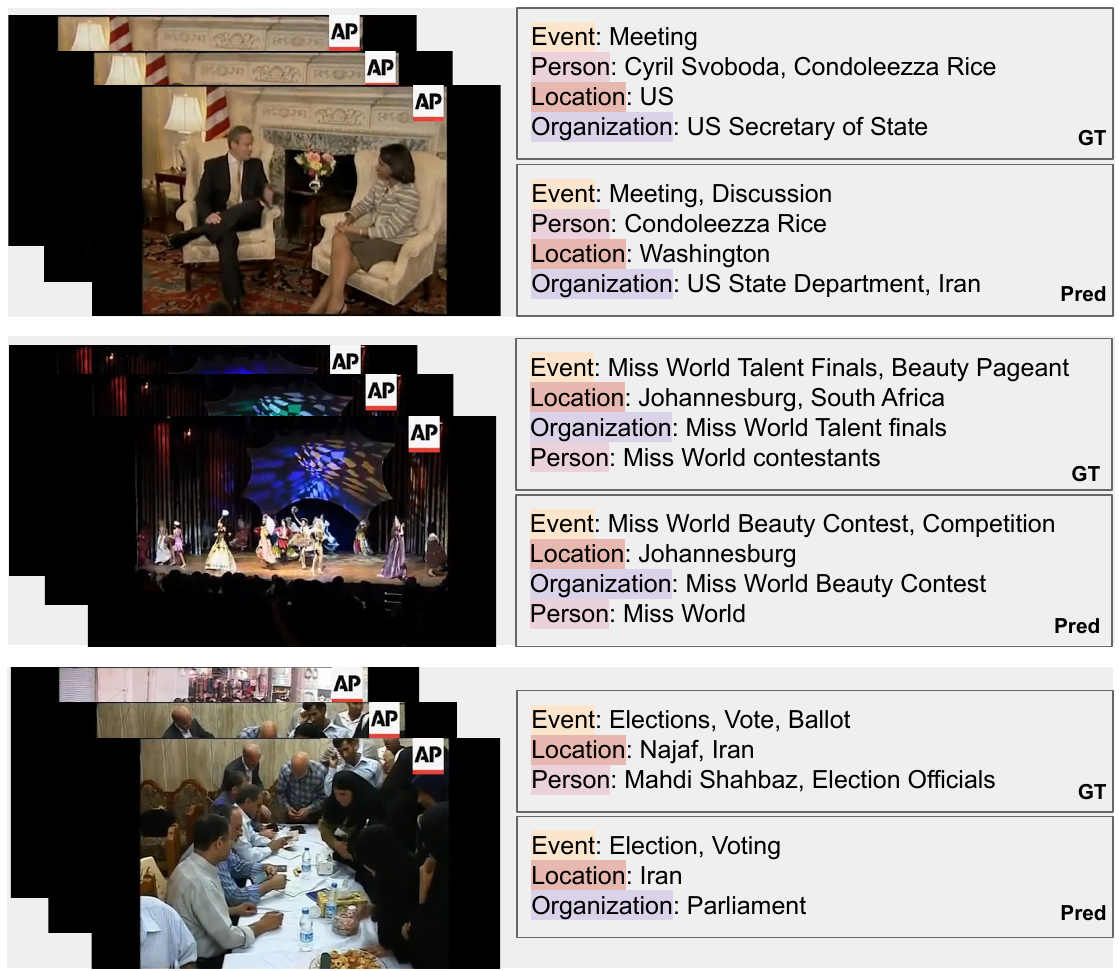}
      \caption{Qualitative analysis of detected entities. Our approach can recognize even obscure entities like Johannesburg, Iran, and so on.}
      \label{fig:entities-comp}
  \end{minipage}
  \begin{minipage}[t]{\linewidth}
    \begin{minipage}[b]{0.28\linewidth}
\centering
\captionsetup{skip=6pt}
\fontsize{9}{12}\selectfont
\setlength{\tabcolsep}{4pt}
\begin{tabular}{l|c}
\toprule
Model    & CIDEr \\
\midrule
CLIP & 0.20 \\
GIT &   \textbf{10.08}  \\
\bottomrule
\end{tabular}
\captionof{table}{Quantitative analysis of detected entities vs CLIP.
}
\label{tab:entity-eval}
    \end{minipage}%
    \hfill
    \begin{minipage}[b]{0.68\linewidth}
        \captionsetup{skip=0pt}
      \includegraphics[width=\linewidth]{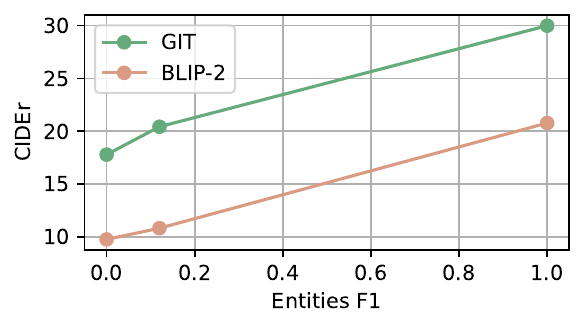}
      \caption{Model performance (CIDEr) improves linearly with entities (F$_1$).
      }
      \label{fig:model-vs-entities}
    \end{minipage}
  \end{minipage}
\end{figure}

\noindent\textbf{Entities are critical.} 
We evaluate the quality of our detected entities in \Cref{tab:entity-eval}. For reference, we use a zero-shot CLIP \cite{radford2021learning} model.
We find that our entities are 50$\times$ better.
Qualitative analysis in \Cref{fig:entities-comp} shows that our approach is quite good in recognizing people (Condoleza Rice in top row), location (Johanessburg and Iran in middle and bottom row respectively), and diverse events (political meeting, cultural event, elections etc). It does this from visual data alone, implying 
that it can learn to recognize places from landmarks and cultural attire, people from a few samples, and organizations from people and locations.

When entities are removed from our pipeline, the model performance drops (see \Cref{tab:ablations-component}). \Cref{fig:model-vs-entities} shows that the model performance is directly proportional to the quality of entities. All this emphasizes the importance of entities.
We further note that there's a large performance gap between detected entity (Ent. F$_1$=10) and ground-truth entity (Ent. F$_1$=1), indicating large returns on investment in entity-perception development. We provide more details and analysis on entities in \cref{sec:ent_blip_pali}.

\begin{figure}[t]
\centering
    \begin{minipage}[b]{0.33\linewidth}
\centering
\captionsetup{skip=7pt}
\fontsize{8}{11}\selectfont
\setlength{\tabcolsep}{3pt}
\begin{tabular}{l l}
\toprule
KE LLM    & BSc\\
\midrule
PaLM-2-340B & \textbf{85.54} \\
PalM-2-24B & 85.50 \\
LlaMA-2-13B  & 85.09 \\
LlaMA-2-7B & 83.56 \\

\bottomrule
\end{tabular}
\captionof{table}{LLMs ablation as KE. Ours is PaLM-2-340B.}
\label{tab:kr-llm-ablation}
  \end{minipage}%
  \hfill
  \begin{minipage}[b]{0.63\linewidth}
  \captionsetup{skip=0pt}
  \centering
  \includegraphics[width=\linewidth]{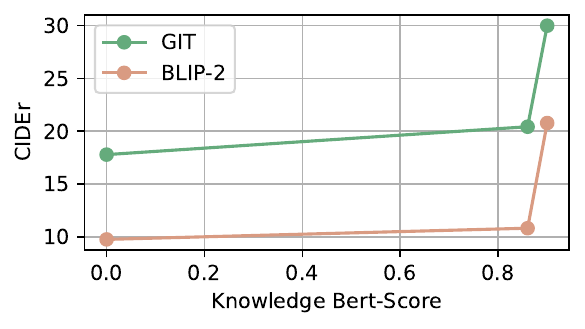}
  \caption{Model performance (CIDEr) improves exponentially with context (BSc).}
  \label{fig:model-vs-knowledge}
  \end{minipage}
\end{figure}

\noindent\textbf{Context matters.}
To assess the quality of our context, we compare it to the context generated by different versions of PaLM-2 and LLaMA-2 \cite{touvron2023llama} as shown in \Cref{tab:kr-llm-ablation}. We measure context quality using BERTScore (BSc), measuring contextual relevance with the ground-truth caption, as traditional metrics (B-4, R-L, and CIDEr) are unsuitable for long paragraph comparisons. We observe that our context scores higher than the others. As expected, a general trend is that larger LLMs achieve higher scores.

The importance of context in our pipeline can be deduced from \Cref{tab:ablations-component}: the performance drops if it's removed. Further, \Cref{fig:model-vs-knowledge} shows that model performance improves exponentially with higher context BSc. This implies that even minor improvements in context quality can drive significant gains in model performance, highlighting a promising area for future research. 



\begin{figure*}[t]
\captionsetup{skip=3pt}
\centering
  \includegraphics[width=\linewidth]{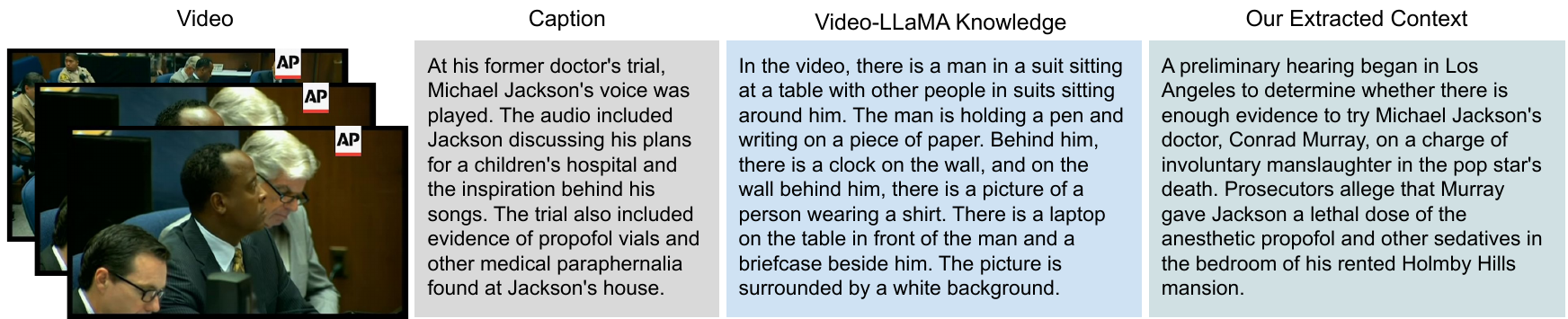}
  \caption{Comparison between our method and a video-based LLM (Video-LLaMA) to extract context. Video-LLaMA fails to extract relevant news knowledge; rather it just describes the video content in a generic manner.}
  \label{fig:knowledge-comp}
\end{figure*}

\begin{table}[t]
\centering
\captionsetup{skip=3pt}
\fontsize{9}{11}\selectfont
\begin{tabular}{l c c}
\toprule
Method    & Bert-Score & Ent. F$_1$\\
\midrule
Single-Stage KE (Video-LLaMA) &   71.11 & 1.24 \\
Two-Stage KE (Ours) & \textbf{85.54} & \textbf{13.53} \\

\bottomrule
\end{tabular}
\caption{Our two-stage KE significantly outperforms Video-LLaMA's single-stage KE, showing that extracting context from detected entities is better than from videos directly.}
\label{tab:knowledge-ours-vs-llama}
\end{table}

\noindent\textbf{Single stage vs our two-stage KE.}
Video-based LLMs, like Video-LLaMA \cite{damonlpsg2023videollama} and VideoChat \cite{li2024videochat}, have proven to be effective in understanding visual content and leveraging related factual knowledge.
As such, we strive to study if a Video-LLaMA based single-stage KE can simply replace our two-stage pipeline of first learning entities to then extract contextual knowledge. To this end, we input the video to Video-LLaMA and task it to extract a relevant news article. 
We observe in \Cref{tab:knowledge-ours-vs-llama} that Video-LLaMA glaringly underperforms our approach despite exhaustive prompt engineering. 

Qualitative comparison of caption, Video-LLaMA context, and our context in \Cref{fig:knowledge-comp} shows that Video-LLaMA keeps trying to describe the video instead of extracting a relevant news information. In comparison, the context extracted by our method is quite good: top row shows that we can extract information about the trial of Michael Jackson's doctor, Conrad Murray, just from the visual data.
This demonstrates that extracting context from learned entities, rather than directly from videos, improves factuality by reducing task complexity. It also implies that, despite the success of video-based LLMs, they still lag behind text-based LLMs on knowledge-intensive tasks like ours.



\begin{figure}[t]
\captionsetup{skip=3pt}
\centering
  \includegraphics[width=\linewidth]{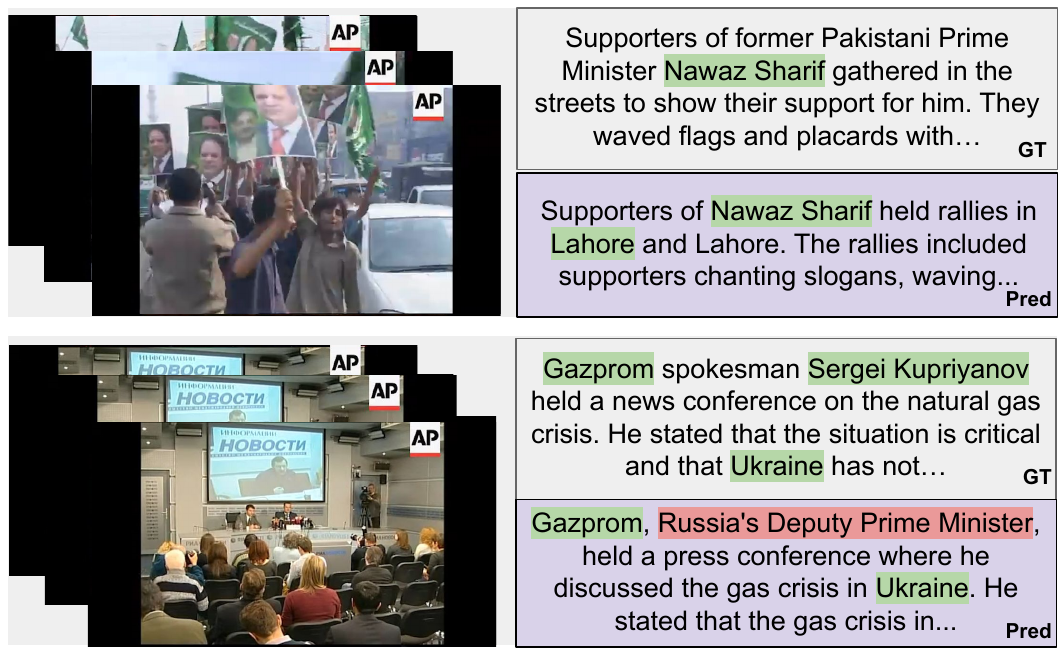}
  \caption{Qualitative results of our Git + \vi model on VIEWS. Green and red highlights denote correctly and wrongly detected entities respectively.}
  \label{fig:qual_analysis}
\end{figure}

\subsection{Qualitative Analysis}
We qualitatively analyse the predictions from our approach in \Cref{fig:qual_analysis}. We see that our method effectively summarizes news video while being informative by predicting relevant named-entities (Nawaz Sharif, Gazprom, Ukraine etc). Remarkably, it uses the contextual information to even detect entities not mentioned in the ground-truth captions: Lahore in row 1 
. While it sometimes predicts a wrong entity (Russia Deputy Prime Minister in row 2), it is able to generally summarize video content well (gas crisis in Ukraine in row 2). These qualitative results further provide evidences of our proposed approach’s effectiveness.

We provide more qualitative samples and analysis of failure cases in \Cref{sec:qual_analysis}.

\subsection{Generalization.}
\label{sec:exp_generalization}

\textbf{Generalization over time.} Our use of an LLM as a plug-and-play Knowledge Extractor (KE) offers unique advantages. 
It allows us to replace our current LLM with a better or more updated one, possessing more recent knowledge or even web search capabilities, while keeping the rest of the model unchanged. 
This enables our approach to generalize over time to novel news content. 
The only caveat is that our frozen CM should adapt to new contexts without re-training.
To test this, we trained the model on data up to January 1, 2017, and the data from after that date acts as new content for evaluation. 
Since our LLM's cut-off date is beyond content date in eval set, it suffices as ``updated" LLM.
The reported results in \Cref{tab:time-gen} demonstrate that our approach can indeed adapt to novel content, and thus generalize over time.

\begin{table}[t]
\centering
\captionsetup{skip=3pt}
\fontsize{9}{11}\selectfont
\setlength{\tabcolsep}{4pt}
\begin{tabular}{l|lll|l}
\toprule
Model  & B-4   & R-L & CIDEr & Ent. F1 \\
\midrule
GIT  & 4.59 & 19.58  & 16.51 & 19.86 \\
\rowcolor{aliceblue}
\hspace{5pt} + $\mathcal{VI}$ \small{(Ours)}  & {4.65} & {19.60} & {18.01}{$_{\green{\Delta} 1.50}$} & {20.72}{$_{\green{\Delta} 0.96}$}\\

\bottomrule
\end{tabular}
\caption{Our method generalizes across time when trained on data before 1 Jan, 2017 and evaluated on data after that date. This 
demonstrates our method's ability to adapt to new context when available.}
\label{tab:time-gen}
\end{table}

\noindent\textbf{Generalization across datasets and modality.} As discussed before, existing video news datasets contain captions containing information from articles as well, which makes them unsuitable for direct video to caption generation. However, we test the limits of our proposed approach on one such dataset, MM-AVS \cite{fu-etal-2021-mm} in \Cref{sec:existing_news_video_results}. We also evaluate our approach on an news image captioning dataset \cite{liu2021visual} in \Cref{sec:existing_news_video_results}. We find that our approach is able to generalize to these new benchmarks as well. 

\section{Conclusion}
\label{sec:conclusion}

In this work, we proposed the new task of generating entity-aware captions from videos only. We also introduced a dataset, VIEWS, to help spur research in this field.    
The task is challenging for existing captioning models as it requires recognizing named-entities in the video and also the context in which the event in video is happening. 
We proposed a modular approach to address these challenges, which we demonstrate improves over vanilla video-only model. 
Still, the relatively modest numbers suggest the task remains challenging and exciting for further research. We provide insights into these future research directions through our rich experiments and ablation studies 
\section*{Limitations}
\label{sec:limitations}

We present a benchmark that challenges models to
detect entities and context from visual cues, and
use them to generate entity-rich video captions. We
assume that there’s a knowledge source out there
that does contain information about this event. We
focus the evaluation of this benchmark to detect
correct entities, extract relevant context, and integrate them into a coherent caption. We do not
cater to videos beyond the scope of this dataset or
daily news videos. While this is a nice property to
have, our evaluation does not prioritize this, but our
method design does allow modular updates with
new KEs to take care of these situations. We leave
this for future work.
\section*{Acknowledgements}
\label{sec:acknowledgements}

We want to thank Junzhang Liu and Venkat Suprabath Bitra for helping us with the human hallucination study.

\bibliography{acl_latex}

\clearpage
\setcounter{page}{1}
\appendix

\section*{Appendix}

\begin{figure*}[t]
\centering
  \includegraphics[width=\linewidth]{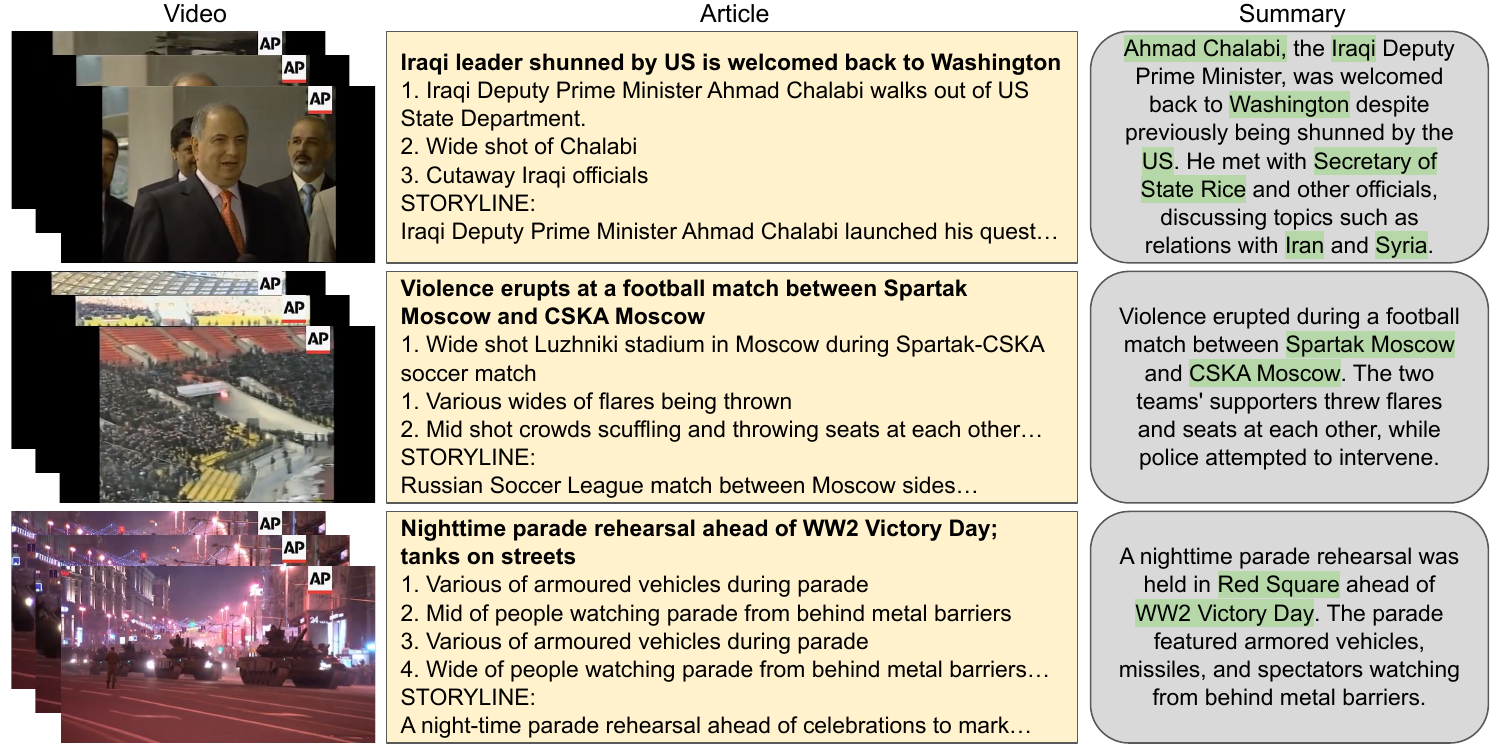}
  \caption{We present more samples from our dataset highlighting key properties: high caption alignment with videos and rich presence of entities (marked in green).}
  \label{fig:dataset_details}
\end{figure*}

\begin{figure}[t]
\centering
    \includegraphics[width=\linewidth]{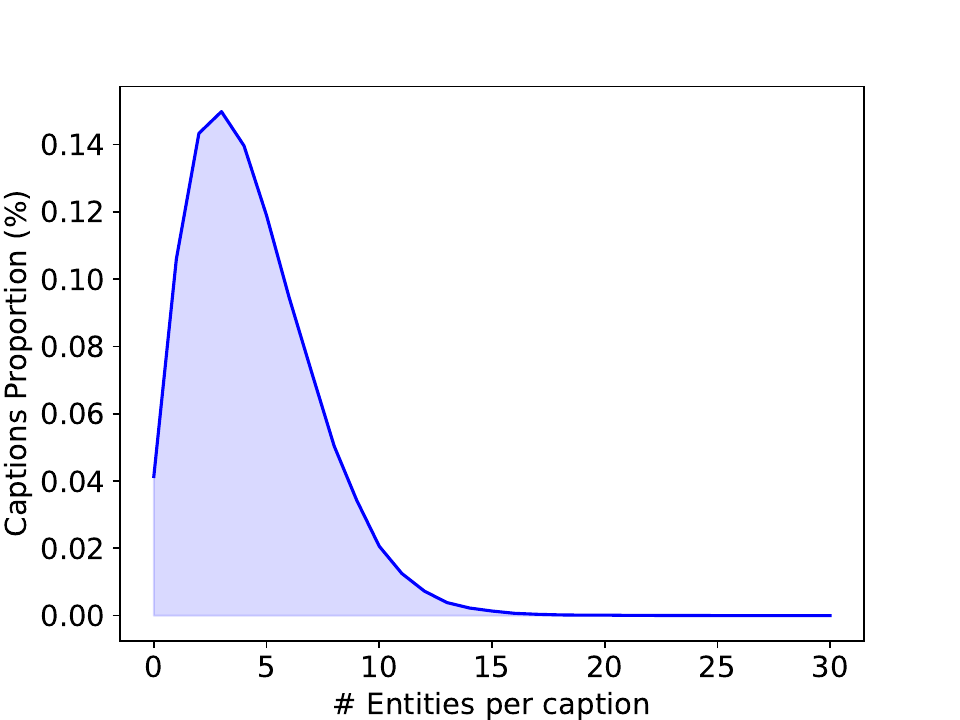}
\caption{Entity distribution per caption.}
    \label{fig:entity_dist_per_cap}
\end{figure}

We provide additional details and analysis to further support the contributions of this work:
\begin{itemize}
    \item Dataset Details (\Cref{sec:dataset_details})
    \item Method Details (\Cref{sec:method_details})
    \item Baselines Details (\Cref{sec:baselines_details})
    \item Effect of Predicted Entity Structure (\Cref{sec:entity_structure})
    \item Effect of utilizing ASR (\Cref{sec:asr})
    \item Qualitative Examples (\Cref{sec:qual_analysis})
    \item Further Analysis of Detected Entities (\Cref{sec:ent_blip_pali})
    \item Generalization across Dataset and Modality (\Cref{sec:existing_news_video_results})
    \item{Case Study on Human Performance on the Dataset}
    (\Cref{sec:case_study})

\end{itemize}

\section{Dataset Details}
\label{sec:dataset_details}

We provide more samples from our dataset in \Cref{fig:dataset_details}. The samples highlight key properties of our dataset:
\begin{itemize}
    \item The bullet summaries used to generate ground-truth captions are highly video-aligned and precise event descriptions of the video.
    \item The ground-truth captions are very faithful to the video content.
    \item The ground-truth captions are rich in entities. \Cref{fig:entity_dist_per_cap} plots the distribution of entities per caption demonstrating that majority of the captions contain around 4-5 entities.
\end{itemize}

\begin{figure*}[t]
\centering
  \includegraphics[width=0.8\linewidth]{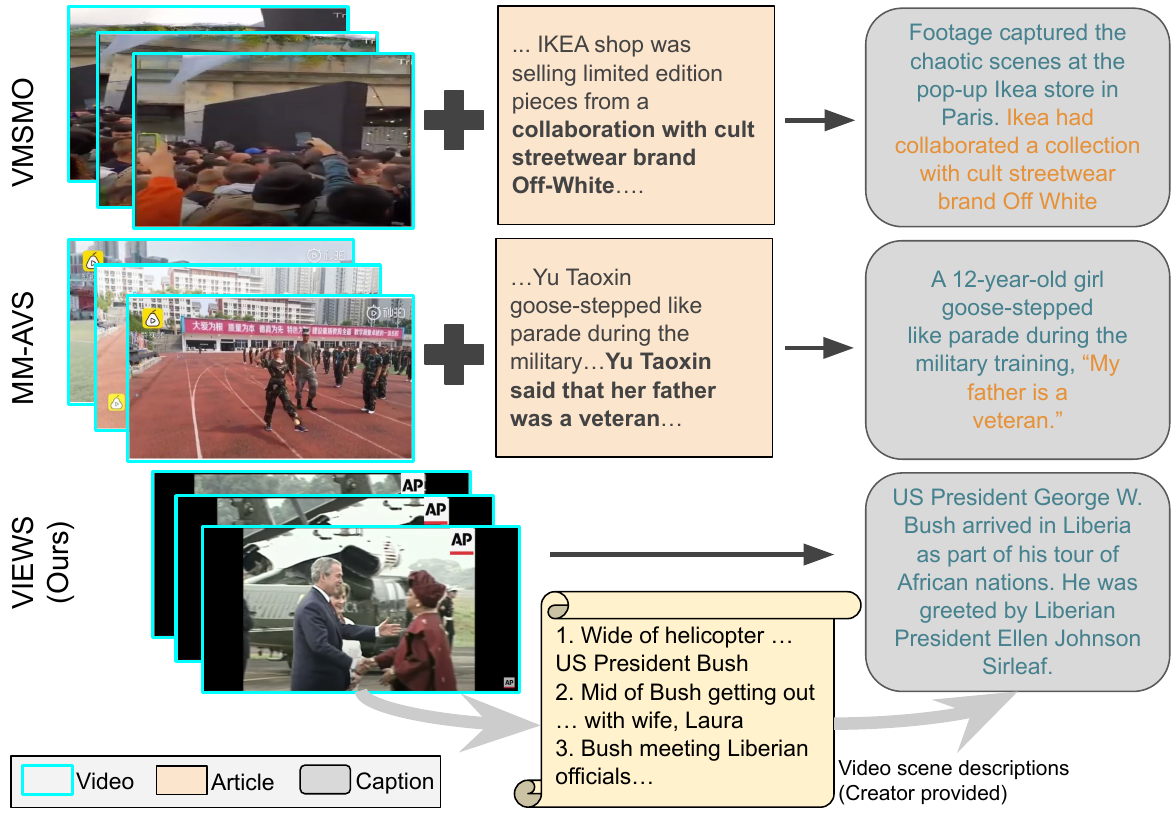}
  \caption{Video-Caption alignment comparison of VIEWS (Ours) against prior News Video Captioning datasets. For each dataset, sentences in {\color{gray} caption} are colored according to their source: {\color{cyan} video} or {\color{orange} article}. MM-AVS (Top) and VMSMO (Middle) both contain sentences in captions that are sourced from articles; thus their captions are not well-aligned with the video. In contrast, all sentences in VIEWS are sourced from video, resulting in high video-caption alignment. The key to this high alignment is generating captions from creator provided video scene descriptions.}
  \label{fig:dataset_comp_details}
\end{figure*}

In \Cref{fig:dataset_comp_details}, we illustrate our dataset's high video-caption alignment compared to prior News Video Captioning datasets (MM-AVS and VMSMO). The source of information for each sentence in the caption is depicted by the color of the sentence. From the figure, we see that captions in both MM-AVS and VMSMO are sourced from videos \textit{and} paired articles. This implies that video is not enough to generate ground-truth captions. In comparison, all sentences in VIEWS's captions are sourced from video. This leads to high video-caption alignment, allowing captions to be generated directly from videos without needing a paired article.
Note that MM-AVS and VMSMO represent the general issue of low video-caption alignment in all existing News Video Captioning datasets.

The reason for high video-caption alignment in our dataset is rooted in our pipeline for generating ground-truth captions. We utilize the creator-provided video scene descriptions, uniquely available in our parent dataset (M$^2$E$^2$R), to generate video captions. This leads to captions whose information source is grounded in videos. This property gives our dataset a distinctive advantage over other news video captioning datasets. The full comparison to prior datasets is presented in Table \ref{tab:additional-dataset-comp}.

\begin{table*}[t]
\captionsetup{skip=3pt}
\centering
\fontsize{8}{10}\selectfont
\setlength{\tabcolsep}{2pt}

\begin{tabular}{l l l c c c c}
\toprule

\multirow{2}*{Dataset}       & \multirow{2}*{Domain}  & \#Videos &  \#Length & \#Words  & Entities & Cap.-Vid \\
&&& Avg. (s) & Avg.  & in Cap. & Align $\uparrow$ \\
\midrule
MSR-VTT~\cite{Xu2016MSRVTTAL}       & Open    & 7.18K      & 20     & 13       & \xmark    & \cmark \\
ActyNet Cap~\cite{krishna2017densecaptioning}   & Open    & 20K       & 180    & 14      & \xmark  & \cmark    \\
YouCook II~\cite{ZhXuCoAAAI18}    & Cooking & 2K        & 316    & -       & \xmark   & \cmark  \\
TVR/TVC~\cite{lei2020tvr}       & TV show & 21.8K     & 76.2   & 13.4    & \xmark *   & \cmark  \\
LSMDC~\cite{rohrbach2016movie}         & Movie   & 128K      & 4.1    & 9        & \xmark *  & \cmark  \\
MM-AVS~\cite{fu-etal-2021-mm}        & News    & 2.17K     & 109    & 57    & \cmark    &  \xmark  \\
XMSMO~\cite{tang2022tldw}         & News    & 4.8K      & 346    & 12      & \cmark  & \xmark  \\
MLASK~\cite{krubinski-pecina-2023-mlask}         & News    & 41.2K     & 86     & 33     & \cmark  &   \xmark  \\
\midrule
VIEWS (Ours)  & News    & 144K       & 148.9 & 46      & \cmark  & \cmark \\

\bottomrule
\end{tabular}

\caption{Additional details on dataset comparison.}
\label{tab:additional-dataset-comp}
\end{table*}

\noindent \textbf{LLM Based Ground-Truth Caption Generation.} The first step in generating ground-truth video captions is to filter out the highly video-aligned and precise event descriptions (bullet summaries) from the paired article. We use simple regex for this purpose. The next step is to generate captions from the bullet summaries. Towards this goal, we utilize PaLM-2-L-IT \cite{anil2023palm} large language model (LLM). We input to the LLM the following prompt:

\begin{lstlisting}[backgroundcolor = \color{backcolor}, basicstyle=\small, breaklines=true]
Task: In this task, you will be given a context and your task is to summarize it in a few sentences. Please don't include specific dates in the summary.

Context: <bullet_summaries>

Instruction: Generate a summary of the Context in around 40 words.
\end{lstlisting}

The resulting captions are highly aligned with the video content as they have been summarized from video event descriptions provided in the paired article. We also specifically remove dates from captions as dates are quite difficult to predict from just the visual signal \cite{fu-etal-2022-theres}.

\vspace{5pt}

\noindent \textbf{LLM Based Caption Quality Control.} While the generated captions are already of high quality, we further control the quality of captions for test and dev set. To this end, we input the captions to LLaMA-2 LLM ask it to rate whether the captions is of high quality or not. We provide the LLM the following prompt for this purpose:

\begin{lstlisting}[backgroundcolor = \color{backcolor}, basicstyle=\small, breaklines=true]
In this task, you will be given a context and its summary. Your job is to determine whether the quality of the summary is high or not. 

A high quality summary has all of the following characteristics:
- It contains the important entities (person, place, organization etc.) in the context 
- It does not contain hallucinations
- It does not leave out critical information from the context

Context: <bullet_summaries>

Summary: <summary>

Instruction: Is Summary of high quality? Please choose between Yes and No.
\end{lstlisting}

As discussed in \cref{sec:dataset}, the captions which are flagged poor is sent for manual verification and correction. This has been done by the authors themselves. In addition to LLaMA-2, we also used PaLM-2 to rate the captions and found similar results.

\section{Method Details}
\label{sec:method_details}

\noindent \textbf{Ground-Truth Entity Extraction.} We detect entities from video using Entity Perceiver (EP). To train EP, we require ground-truth entities. This is generated by extracting entities from bullet summaries using a LLM (PaLM-2-L-IT). We use a LLM for extracting these ground-truth entities as they have shown to excel in these kinds of Information Extraction (IE) Named-Entity Recognition (NER) tasks \cite{wang2023gptner}. 
To this end, we input the following prompt to the LLM:

\begin{lstlisting}[backgroundcolor = \color{backcolor}, basicstyle=\small, breaklines=true]
Task: In this task, you will be provided a piece of text and your task would be to extract named entities from the text. Please don't include any date entity.

Text: <bullet_summaries>

Instruction: Extract named entities from the following text in this format: {<Entity_Type>: [<Entity_list>]}.
\end{lstlisting}

Again, we don't extract any date entity as they are difficult to predict from visual data.

\vspace{5pt}
\noindent \textbf{Knowledge Extraction (KE).} Once we predict entities using KP, we utilize them to extract a relevant news article as a knowledge source. We again use the same LLM as KE for this purpose. The prompt given is:

\begin{lstlisting}[backgroundcolor = \color{backcolor}, basicstyle=\small, breaklines=true]
Task: In this task, you will be provided a list of entities and your task would be to retrieve a summary of the most relevant news article related to those entites.

Entities: <entities>

Instruction: Retrieve a summary of the most relevant news article involving these Entities.
\end{lstlisting}

While use of online web search may seem a natural choice for KE, they have certain limitations. They are sensitive to search query and can be quite noisy. Besides, only limited search APIs are avaiable publicly. For example, Google search charges for $>100$ queries/day\footnote{\url{https://developers.google.com/custom-search/v1/overview?hl=ja}}.

\section{Baselines Details}
\label{sec:baselines_details}

\noindent \textbf{GIT}\cite{wang2022git} is a state-of-the-art vision-language model that has demonstrated strong results on 12 benchmarks by scaling up data and model size. 
To finetune it on our task, we follow the proposed approach of concatenating all frames' patch embeddings together and feeding them to the model to generate captions. 
We also concatenate frames' tokens with \vi tokens before inputting to the model. 
We use the official pretrained GIT-L with ViT-L/14 image encoder\cite{dosovitskiy2020image} and 6 layer transformer decoder \cite{vaswani2017attention}.

\noindent \textbf{BLIP-2}\cite{li2023blip2} has shown strong results on many knowledge-intensive tasks by leveraging a frozen LLM in its architecture.
They utilize a Q-Former \cite{li2023blip2} to bridge the embedding space between visual signals and textual space of a LLM.
During training, only Q-Former is learned while keeping LLM and the image encoder frozen.
We also finetune only the Q-Former as in the original implementation.
We utilize the Q-Former to extract queries for each frames, concatenate them with temporal embeddings, and then join them together before feeding into the LLM. 
Again, for our pipeline, these query embeddings are also concatenated with context tokens before inputting to LLM. 
We finetune the official BLIP-2 pretrained weights obtained using ViT-L/14 image encoder and Flan-T5-XL LLM \cite{chung2022scaling}.

\noindent \textbf{ViT-T5.} We construct this baseline to leverage the best of both LLMs and fine-tuning. To this end, ViT-g/14 \cite{fang2023eva} is employed as the image encoder and T5-L \cite{chung2022scaling} as the LLM decoder. Per frame Q-Former is utilized, as in BLIP-2, to extract query embeddings and appended with temporal embeddings. These query embeddings together with our entity and context tokens are concatenated together to form the input to the LLM decoder. We finetune it end-to-end. 
The model is pretrained on WebLI data \cite{chen2022pali} and scores a strong CIDEr \cite{vedantam2015cider} of 69.2 on MSRVTT Video Captioning benchmark when finetuned.

\section{Effect of Predicted Entity Structure}
\label{sec:entity_structure}

We predict entities in a structured format as shown in \cref{fig:entities-comp}. This structure is in the form of a dictionary and follows this format: 
\begin{equation}
    \{<Entity\_Type>: [<Entity\_List>]\}  \nonumber
\end{equation}

The structure relates each entity to its type. This information is important in the downstream task of Knowledge Extraction (KE). For example, in the case of a United Nations (UN) summit held at, say, Geneva, the United States (US) acts as an organization not a location. Not delineating the entity type of United States may give the false impression that the UN summit was held in the US. We measure the quantitative impact of this structured entity detection in \cref{tab:knowledge-entities-wo-structure}. We find that there is marked degradation in the quality of extracted knowledge (Ent. F$_1$ 13.53 to 12.32), if we don't input entities in the structured format. This demonstrates the usefulness of detecting entities in the aforementioned format.

\begin{table}[t]
\centering
\begin{tabular}{l c c}
\toprule
Model    & Bert-Score & Ent. F$_1$\\
\midrule
Ours &   \textbf{85.54} & \textbf{13.53} \\
\hspace{5pt} - w/o Entities Structure & 85.47 & 12.32 \\

\bottomrule
\end{tabular}
\caption{Comparison of extracted knowledge with and without entity structure.}
\label{tab:knowledge-entities-wo-structure}
\end{table}
\begin{table}[t]
\centering
\begin{tabular}{l|ccc}
\toprule
Model    & BLEU-4   & Rouge-L & CIDEr  \\
\midrule
GIT &  5.41 & 21.88 & 17.78 \\
\hspace{5pt} + ASR &  5.42 & 21.91 & 17.84 \\
\midrule
GIT + Context &  \textbf{5.62}  &  \textbf{22.12}  &  \textbf{21.0}  \\
\hspace{5pt} + ASR &  5.57 & 22.0 & 20.22 \\

\bottomrule
\end{tabular}
\caption{Effect of using ASR on our task.}
\label{tab:asr-ablations}
\end{table}

\section{Effect of utilizing ASR}
\label{sec:asr}

\begin{figure*}[t]
\centering
  \includegraphics[width=0.9\linewidth]{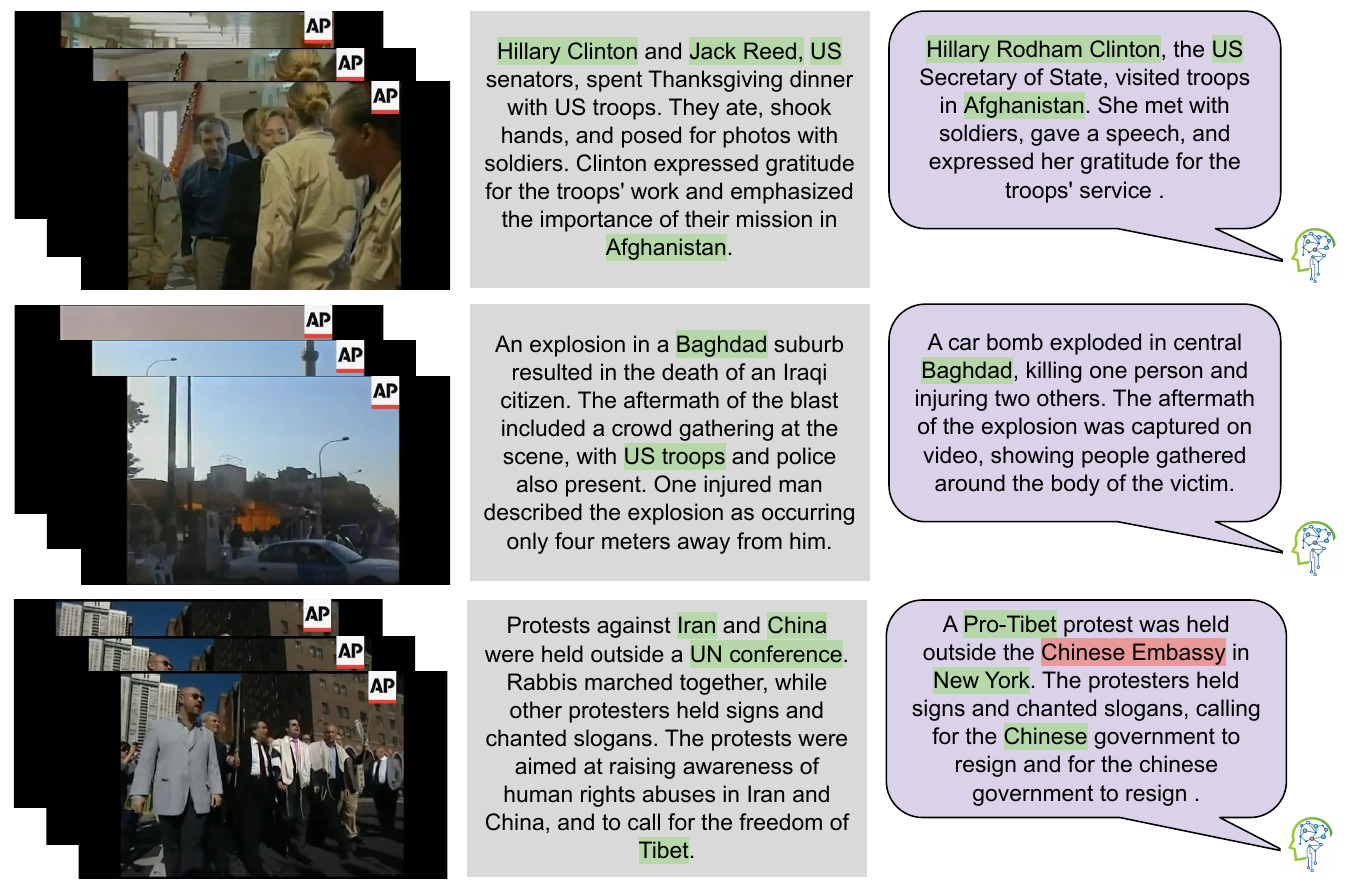}
  \caption{Additional sample predictions from our approach VIEWS. Left: Video; Middle: Ground-Truth Captions; Right: Prediction. Green highlights denote correctly detected entities, while red highlights denote wrongly predicted entities.}
  \label{fig:qual_analysis_appendix}
\end{figure*}

\begin{figure*}[t]
\centering
  \includegraphics[width=\linewidth]{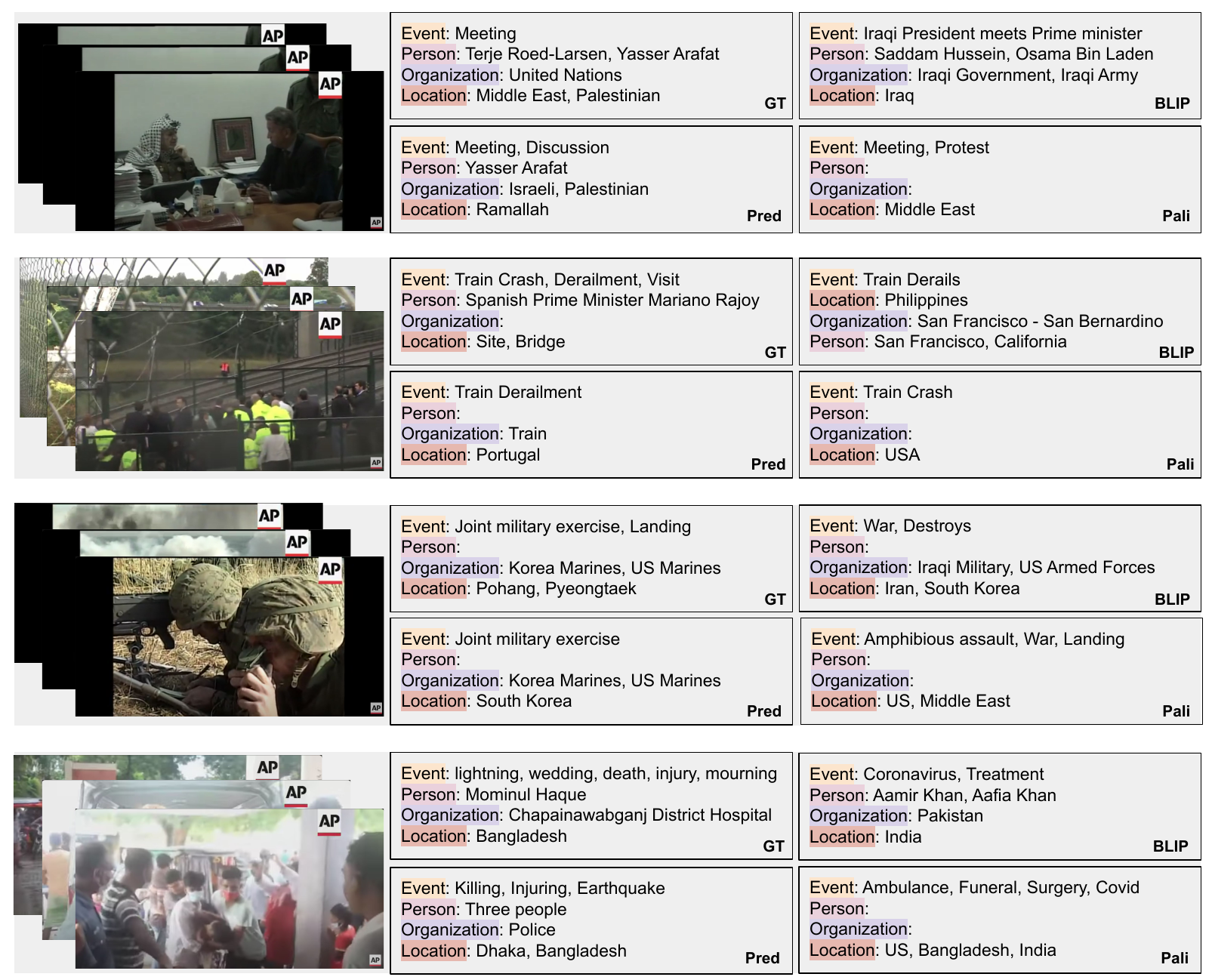}
  \caption{Qualitative comparison of our Entity Detection module against BLIP (BLIP-2) and Pali (Pali-X).}
  \label{fig:entities_blip_pali}
\end{figure*}

Automatic Speech Recognition (ASR) is an important tool in video captioning. It captures audio signals from the video in text form and provides this information as complementary to visual signals. This information is especially impactful in summarizing instructional videos \cite{yang2023vid2seq}. It would also have been beneficial in a narrative or reportive news video. However, most videos in our dataset is descriptive, in the sense that it just captures the news event video without any narrator or reporter. This is exemplified by the fact that only 40\% of our data contains any ASR content. To further illustrate the insignificant impact of ASR on our dataset, we compare our models with a version trained with ASR signals. We report the results in \cref{tab:asr-ablations}. We see that adding ASR to GIT barely improves the performance (17.84 vs 17.78 CIDEr). This is far below the improvement recorded by our apporach which adds entities and external knowledge as Context (21.0 vs 17.78 CIDEr). We further note that adding ASR to our Context depreciates model's performance a bit (21.0 to 20.22 CIDEr). We hypothesize that this may be due to the fact that our approach already provides relevant context to the model and adding ASR may just create noise. These results highlight that while ASR may be useful for some video captioning tasks, it's not very relevant to our news summarization task.

\section{Qualitative Examples}
\label{sec:qual_analysis}

We provide additional qualitative examples in \cref{fig:qual_analysis_appendix}. These examples further illustrate our approach's effectiveness in generating entity-aware captions. The model is able to detect people (Hillary Clinton in row 1), place (Afghanistan in row 1 and Baghdad in row 2) and even detect novel entity not present in ground-truth (New York in row 3). Further, the model is able to effectively integrate these entities into informative captions.

\noindent\textbf{Failure Analysis.} While our method generally performs well, we seek to analyze its failure cases to ascertain its limitations. In \Cref{fig:qual_analysis}, we find that our model is unable to detect Gazprom spokesman Sergei Kupriyanov, who is not a very prominent person. This follows a general pattern that entities on the long tail are, at times, wrongly predicted. Further, we see in \Cref{fig:qual_analysis_appendix} that the model wrongly mistakes UN Conference for Chinese Embassy. This is another common mistake wherein the model fails to detect entities that have non-specific visual details in the video.

\section{Further Analysis of Detected Entities}
\label{sec:ent_blip_pali}

To detect named-entities from video, we train an Entity Perceiver (EP). In our implementation, EP is a GIT model trained on ground-truth entities extracted from video bullet summary using a LLM. We analysed the performance of EP in \cref{tab:entity-eval} and \cref{fig:entities-comp}. In this section, we compare EP against state-of-the-art LLM based Vision-Language models (VLMs): BLIP-2 \cite{li2023blip2} and PaLI-X \cite{chen2023palix}. These models have demonstrated strong zero-shot results on a number of vision-language benchmarks. As such, we use them to detect entities and compare how our proposed EP fares against them. 

To do so, we first sample representative frames from the video. Next, we input them to BLIP-2 and PaLI-X and prompt it predict the salient event, person, location and depicted organization in the video. A union of predictions from all the frames is collected as the final prediction. We compare the results against EP in \cref{fig:entities_blip_pali}. Our initial exploration with these VLMs returned poor results. As such, we just demonstrate the performance comparison with our EP qualitatively. BLIP-2 is bullish in its predictions, wrongly detecting person Saddam Hussein (row 1) and Aamir Khan (row 4), and getting location and organization wrong in almost all samples. On the other hand, PaLI-X makes much less predictions and in the process doesn't predict person or organization in any sample. In contrast, EP correctly detects a number of named entities (Yasser Arafat, row 1; South Korea, US and Korean marnies, row 3; Bangladesh, row 4) while making much lesser false predictions. This highlights the superior performance of our trained EP in comparison to existing SOTA Vision-Language models.

\begin{figure}[t]
\centering
    \includegraphics[width=\linewidth]{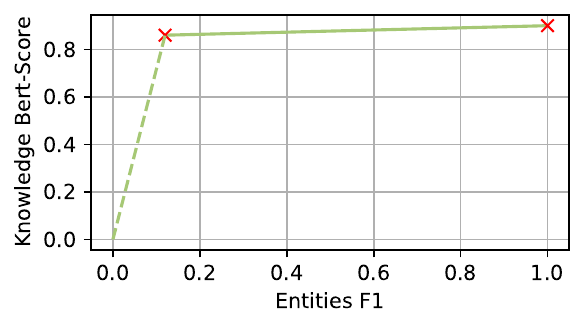}
    \caption{Impact of entities on extracted contextual knowledge.}
    \label{fig:knowledge_vs_entities}
\end{figure}

Additionally, we also quantitatively analyse GIT's performance as Entity Perceiver. To this end, we compare GIT's output to a reference baseline constructed using CLIP \cite{radford2021learning}. To build this reference baseline, we query video-frames to a pretrained CLIP to retrieve the closest entity from train-set entities. Entities retrieved for all the frames are joined together to form the final predictions. We sample 6 frames for each video for this baseline. 
We report the results on validation set in \Cref{tab:entity-eval-clip}. From the table, we see that GIT outputs entities that are clearly better than CLIP. The result provides further evidence of high quality entities detected by our EP (GIT).

We also analyse how entities quality impacts contextual knowledge and  in fig \ref{fig:knowledge_vs_entities}. The figure shows that quality of context improves with the entities generally, although it plateaus towards the end. This implies that we may need to make a lot of improvement in entity detection to see a significant improvement in the quality of context.

\begin{table}[t]
\centering
\fontsize{9}{11}\selectfont
\begin{tabular}{l|c|ccc}
\toprule

Model    & Ent. F$_1$  & B-4   & R-L \\
\midrule
CLIP & 0.02 & 1.14 & 16.30 \\
GIT &   \textbf{11.71} & \textbf{27.22} & \textbf{51.86}\\

\bottomrule
\end{tabular}
\captionof{table}{Quantitative comparison of detected entities against reference baseline -- CLIP. GIT outperforms CLIP by a fair margin. 
}
\label{tab:entity-eval-clip}
\end{table}
\begin{table}[t]
\centering
\captionsetup{skip=3pt}
\fontsize{9}{11}\selectfont
\begin{tabular}{lcc}
\toprule





Model & Human $\downarrow $ & GPT-4 $\downarrow$    \\
\midrule
GIT & 64 &  79.2   \\

\hspace{5pt} + $\mathcal{VI}$ \small{(Ours)} & \textbf{60} &  \textbf{78.6} \\

\bottomrule
\end{tabular}
\captionof{table}{Entity Hallucination Study. Lower is better. Our model consistently hallucinates less in both human and GPT-4 evaluation.}
\label{tab:entity-hallucination-study}
\end{table}

\noindent\textbf{Hallucination Study.} While Ent. F$_1$ provides some measure of hallucinated entities in the generated caption, we seek to explicitly quantify entities' hallucination to better understand the model's behavior. To this end, we measure hallucinated in predicted captions via a manual evaluation and a GPT-4 \cite{openai2024gpt4technicalreport} evaluation.

For human evaluation, two expert annotators not associated with this work are asked to detect if the caption predicted by the model contains hallucinated entities. To detect hallucination, we use strict criteria that even if there is a single entity in the predicted caption that is not present in the ground-truth caption, the predicted caption should be marked as hallucinated. A total of 100 random samples are evaluated. The results are reported in \Cref{tab:entity-hallucination-study}. We find that our model hallucinates less as compared to the baseline. Further, the strict hallucination criteria leads to a relatively high hallucination percentage for both the baseline and our model.

For GPT-4 evaluation, we input the ground-truth caption and the predicted caption to GPT-4 and task to detect it to detect hallucination in the predicted caption. Again, we use the strict criteria that even a single entity in the prediction caption that is not present in the ground-truth would make it hallucinated. Even for this evaluation, we find from \Cref{tab:entity-hallucination-study} that our model performs better than the baseline. A total of 500 random samples were used for this evaluation.

\section{Generalization across Dataset and Modality}
\label{sec:existing_news_video_results}

\noindent \textbf{Generalization across Dataset} As mentioned before, existing news video captioning datasets are too small and the captions are not well aligned with the video content. However, to test the limits of our approach, we also report results on one of them -- MM-AVS. Other datasets are not publicly available \cite{tang2022tldw}, not accessible \cite{li-etal-2020-vmsmo}, or not in English \cite{krubinski-pecina-2023-mlask, li-etal-2020-vmsmo}. MM-AVS contains data from CNN and Daily Mail and has a total of 2173 samples. We report our results in \cref{tab:mm-avs}. As can be seen, our approach leveraging context outperforms vanilla video only captioning by 4 CIDEr points. This illustrates the generalization of our approach to other news video datasets. 

\noindent \textbf{Generalization across Modality.} We seek to understand if our approach can generalize to News Image Captioning as well. As such, we evaluate our approach on the largest news image captioning dataset: Visual News. We see \cref{tab:visual-news} that our approach improves our image-only training in this image domain as well. In fact, our approach using context comes close to the performance of some existing approaches that additionally use paired articles (greyed rows) to generate entity-aware captions.

\begin{table}[t]
\centering
\resizebox{1.\linewidth}{!}{
\setlength{\tabcolsep}{2pt}
\begin{tabular}{l|ccc}

\toprule
Model    & BLEU-4   & Rouge-L & CIDEr \\
\midrule
\color{gray}M$^2$SM \cite{fu-etal-2021-mm} & \color{gray} - & \color{gray} 30.23 & \color{gray} -\\
\color{gray}SCCS \cite{qiu2022semanticsconsistent} & \color{gray}- & \color{gray}34.21 &\color{gray}- \\
\midrule
GIT &  7.42 & 16.34 & 51.53 \\
\hspace{5pt}+ $\mathcal{VI}$ (Ours) & \textbf{7.91} & \textbf{16.55} & \textbf{55.83}\\

\bottomrule
\end{tabular}
}
\caption{Evaluation on MM-AVS, a video news captioning dataset. Greyed rows use paired news aritcle as input also.}
\label{tab:mm-avs}
\end{table}
\begin{table}[t]
\centering
\setlength{\tabcolsep}{2pt}
\resizebox{1.\linewidth}{!}{
\begin{tabular}{l|c|ccc}

\toprule
Model    & Ent. F$_1$  & BLEU-4   & Rouge-L & CIDEr \\
\midrule
\color{gray}Tough-to-Beat \cite{biten2019good} & \color{gray}4.9 & \color{gray}1.7 & \color{gray}13.2 & \color{gray} 12.4\\
\color{gray}Pooled Embeddings \cite{biten2019good} & \color{gray}5.3 & \color{gray}2.1 &\color{gray}13.5 &\color{gray} 13.2 \\
\color{gray}Visual News \cite{liu2021visual} &  \color{gray} 18.59 & \color{gray} 5.3 & \color{gray} 17.9 & \color{gray} 50.5 \\
\color{gray}Tell \cite{zhou-etal-2022-focus} &  \color{gray} 23.44 & \color{gray} 11.6 & \color{gray} 25.0 & \color{gray} 107.6\\
\midrule
GIT & 6.9& 5.19 & 15.63 & 26.82 \\
\hspace{5pt}+ $\mathcal{VI}$ (Ours) & 6.9 & \textbf{5.24} & \textbf{15.78} & \textbf{27.93}\\

\bottomrule
\end{tabular}
}
\caption{Evaluation on Visual News, an image news captioning dataset. Greyed rows use paired news article as input also.}
\label{tab:visual-news}
\end{table}

\section{Case Study on Human Performance on the Dataset}
\label{sec:case_study}

We present a challenging task in this work: generating entity-aware captions directly from the videos. To validate that this task is indeed practical, we do a case study where we ask human annotators to write entity-aware captions given videos. The annotators are allowed to use any external source: web, Wikipedia, LLMs etc. The annotators are technical people not associated with this work. The case study is visualized in \Cref{fig:human_case_study}.

From this case study, we make the following observations:
\begin{itemize}
    \item In general, humans can write good entity-aware captions using external sources. Hence, the task is practical.
\begin{figure*}[h!]
\centering
  \includegraphics[width=\linewidth]{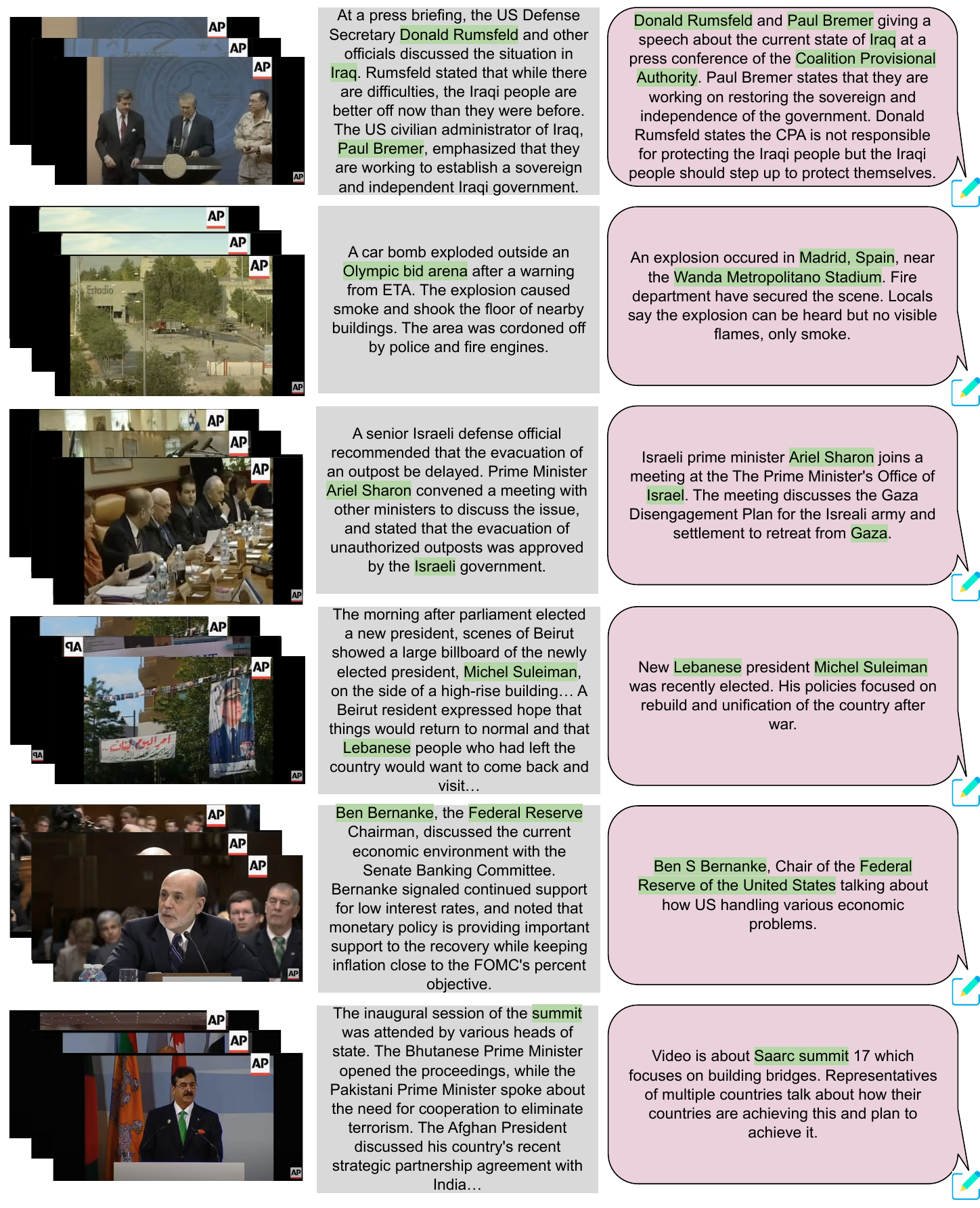}
  \caption{Case study on random video samples to determine the practicality of the task. In general, humans are able to write reasonable entity-aware captions.}
  \label{fig:human_case_study}
\end{figure*}
    \item Humans are able to detect all entities from ground-truth.
    \item The context of the annotated captions are quite good and aligns strongly with the ground-truth caption.
    \item At times, humans can detect entities that are not even mentioned in the ground-truth. For example Coalition Provisional Authority and Wanda Metropolitano Stadium in row 1 and row 2 respectively.
\end{itemize}

From these observations, it's safe to conclude that while the task is challenging, it's practical.

\end{document}